\documentclass{article}

\usepackage{iclr2027_conference,times}

\usepackage{amsmath,amsfonts,bm}

\def\eqref#1{equation~\ref{#1}}

\def\1{\bm{1}}

\DeclareMathAlphabet{\mathsfit}{\encodingdefault}{\sfdefault}{m}{sl}
\SetMathAlphabet{\mathsfit}{bold}{\encodingdefault}{\sfdefault}{bx}{n}

\usepackage{hyperref}
\usepackage{url}
\usepackage[table]{xcolor}
\usepackage{booktabs}
\usepackage{multirow}
\usepackage{float}
\usepackage{graphicx}
\usepackage{enumitem}
\usepackage{amssymb}
\definecolor{bestblue}{RGB}{217,232,252}
\definecolor{secondgreen}{RGB}{221,242,214}

\title{
FORGE: Fault-Oriented Rework Code Generation
for Executable Interlocking Assembly
}

\author{
\begin{minipage}[t]{\dimexpr\textwidth-2\tabcolsep\relax}
\centering
\bfseries
Fuqiang Wang\(^{2,3,4}\),
Jiahao Fu\(^{2,3,4}\),
Ren Hefeng\(^{2,3,4}\),
Yining Qian\(^{5}\),\\
Chuanting Zhang\(^{6}\),
Cheng Tan\(^{7}\),
Yuwen Liu\(^{2,3,4}\),
Jingxuan Wei\(^{1,2,3,4}\)\thanks{Corresponding author.}\\[6pt]
\normalfont
\(^1\) University of Chinese Academy of Sciences, Beijing, China\\
\(^2\) Key Laboratory of Computing Power Network and Information Security,\\
Ministry of Education, Shandong Computer Science Center\\
(National Supercomputer Center in Jinan),\\
Qilu University of Technology (Shandong Academy of Sciences),\\
Jinan 250014, China\\
\(^3\) Shandong Provincial Key Laboratory of Industrial Network\\
and Information System Security,\\
Shandong Fundamental Research Center for Computer Science,\\
Jinan 250014, China\\
\(^4\) Quan Cheng Laboratory, Jinan, China\\
\(^5\) Northeastern University, China\\
\(^6\) Shandong University, China\\
\(^7\) Tencent, China\\[4pt]
{\small\texttt{weijingxuan20@mails.ucas.ac.cn,
10431250134@stu.qlu.edu.cn}}
\end{minipage}
}

\iclrfinalcopy

\begin{document}

\maketitle
\lhead{Preprint}


\begin{abstract}

Large language models have demonstrated strong capabilities in structured
reasoning and code generation, yet identifying an assembly fault does not
guarantee the ability to generate executable rework programs.
In interlocking assemblies, a locally plausible repair program may violate
blocking dependencies, omit required recovery operations, or terminate in an
incorrect configuration.
We identify this challenge as \emph{fault-conditioned executable rework
generation}, where models must jointly reason about structural failures and
generate programs that restore constrained assembly states.
To benchmark this capability, we introduce \textbf{FORGE-Bench}, containing
4,000 tasks across 26 interlocking assemblies with five assemblies held out for
evaluating generalization to unseen structures.
We further propose \textbf{FORGE} (\textbf{F}ault-\textbf{O}riented
\textbf{R}ework Code \textbf{G}eneration for \textbf{E}xecutable Interlocking
Assembly), a structured state-to-program framework that decomposes rework
generation into \textbf{Rework Structuring} and \textbf{Action Grounding}.
Rework Structuring captures fault-induced recovery dependencies, while Action
Grounding maps these dependencies into state-conditioned executable actions.
Experiments show that locally reasonable repair code is insufficient for
reliable assembly recovery, and that connecting structural recovery reasoning
with executable action realization is essential for verified target
restoration.

\end{abstract}

\section{Introduction}
\label{sec:introduction}

Recent large language models are increasingly extending code generation beyond
software development toward structured programs for interacting with complex
environments.
Recent systems have demonstrated strong capabilities in structured generation,
tool use, code synthesis, and long-horizon planning by translating high-level
instructions into machine-readable representations and executable actions
~\citep{zhang2025xlam,liu2025toolace,lei2025planningdriven,zheng2026ebench}.
These advances make language models increasingly attractive for physical
reasoning tasks, where symbolic states, discrete action interfaces, and
structural constraints can be represented in a structured form
~\citep{kim2024openvla,brohan2025pi0,shukor2025smolvla,nvidia2025grootn1,chen2026pointact}.
However, generating a program that appears reasonable does not guarantee that
it remains executable when successful execution depends on hidden dependencies
among multiple operations.

Interlocking assemblies provide a setting where this limitation becomes
explicit.
A faulty component is rarely independent of the surrounding structure.
Repairing it may require removing blocking parts, releasing locking relations,
and restoring temporarily displaced components after the correction.
Therefore, a valid rework program must reason not only about the faulty part,
but also about the dependency chain that determines when each operation can be
executed and how the final assembly state can be restored.
Recent assembly-oriented systems have advanced instruction generation,
hierarchical scene representation, error correction, and context-aware
assistance~\citep{yan2025neuralassembler,li2025vlmmsgraph,wan2024scanet,
liu2025oneshotassembly,wang2026vlabot,wang2026assemblymate}.
Nevertheless, these methods primarily address forward assembly, operator
guidance, or local correction rather than complete dependency-aware recovery.

We call this regime \emph{executable rework generation}, where models must move
beyond isolated fault identification and local repair actions toward
dependency-aware programs that execute complete recovery procedures.
This regime sits at the intersection of assembly reasoning, program synthesis,
and symbolic planning, but none of these directions directly captures the
coupled requirements of fault diagnosis, structural dependency reasoning, and
target-state restoration.
Assembly reasoning methods primarily focus on component relations, assembly
sequences, or forward construction, but do not address recovery from faulty
states~\citep{pun2025brickgpt,wang2026vlabot,wang2026assemblymate}.
Program generation methods study executable code synthesis, yet typically
assume that required operations are already specified or that individual
actions can be evaluated independently
~\citep{chi2025instructflow,ahn2025reliablecap,chen2025codesymbolicplanner}.
Symbolic planning approaches provide constraint-aware reasoning, but rarely
consider the joint problem of identifying faults and generating
state-restoring programs under physical assembly dependencies
~\citep{curtis2025proc3s,petruzzellis2025hvr,pesjak2026robotplanning,
huang2025planningdark,chen2024autotamp}.


\begin{figure}[t]
    \centering
    \includegraphics[width=\linewidth]{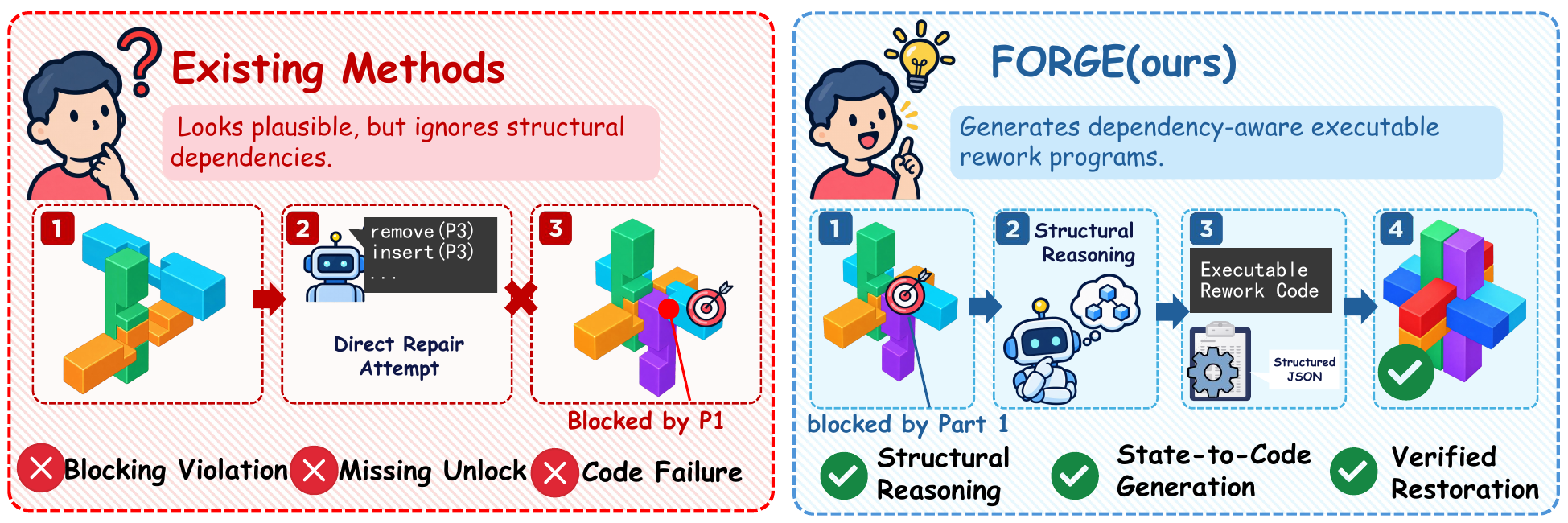}
    \caption{
    Motivation of FORGE.
    Existing methods overlook structural dependencies during assembly rework,
    while FORGE enables executable recovery through dependency-aware reasoning.
    }
    \label{fig:motivation}
\end{figure}

To study this gap, we introduce \textbf{FORGE}, short for
\textbf{F}ault-\textbf{O}riented \textbf{R}ework Code
\textbf{G}eneration for \textbf{E}xecutable Interlocking Assembly.
FORGE formulates interlocking assembly repair as a structured
state-to-program generation framework built on vanilla SFT with two
complementary algorithmic modules: \textbf{Rework Structuring}, which captures
fault-induced recovery obligations among components, and \textbf{Action
Grounding}, which maps these obligations into executable actions, parameters,
and symbolic state transitions.
Given public assembly states and structural constraints, FORGE generates
complete recovery programs while avoiding unnecessary intervention on
assemblies that are already correct.
Unlike benchmarks centered on isolated function calls, FORGE evaluates
state-changing action sequences and their final verified outcomes, reflecting
the growing emphasis on stateful and multi-step agent evaluation.
~\citep{patil2025bfcl,lu2025toolsandbox,chen2025acebench,
zheng2026ebench,shen2026sciagentgym}.
Our main contributions are as follows:
\begin{itemize}[leftmargin=2em]
    \vspace{-1mm}
    \item We identify and formalize \emph{fault-conditioned executable rework generation}, a structured task requiring joint fault diagnosis, dependency-aware recovery planning, executable action generation, and target-state verification.

    \vspace{-1mm}
    \item We introduce \textsc{FORGE-Bench}, a benchmark for interlocking-assembly
rework with assembly-disjoint splits, controlled fault/no-fault cases, and
verifier-based evaluation of structural correctness and target-state
restoration.

    \vspace{-1mm}
    \item We propose \textsc{FORGE}, a dependency-aware framework that separates recovery into \textbf{Rework Structuring} and \textbf{Action Grounding}, enabling state-conditioned generation of executable restoration programs.
\end{itemize}
\section{Related Work}
\label{sec:related_work}

\textbf{Assembly reasoning.}
Assembly reasoning focuses on representing component relations, spatial
dependencies, and executable operations in physical environments.
General-purpose vision-language-action models such as OpenVLA, $\pi_0$,
SmolVLA, and GR00T N1 connect language understanding with robotic control,
while OpenVLA-OFT and PointACT improve fine-tuning efficiency and
spatially grounded manipulation
~\citep{kim2024openvla,brohan2025pi0,shukor2025smolvla,nvidia2025grootn1,
kim2025openvlaoft,chen2026pointact}.
Assembly-oriented methods further study instruction generation, hierarchical
scene representations, error correction, and physically stable construction
~\citep{yan2025neuralassembler,li2025vlmmsgraph,wan2024scanet,pun2025brickgpt}.
Recent manufacturing systems also explore autonomous assembly, VLA-mediated
robotic manipulation, and multimodal operator assistance
~\citep{liu2025oneshotassembly,wang2026vlabot,wang2026assemblymate}.
Together, these studies provide useful representations of components and
relations, but a repair procedure must also account for parts temporarily
removed to access a fault.
Existing approaches mainly address forward construction, local correction,
or instruction following.
FORGE instead targets faulty interlocking states and requires joint fault
diagnosis, dependency-aware rework, and target-state restoration.

\textbf{Executable planning.}
Executable planning converts high-level goals into action sequences or
programs that satisfy environmental constraints.
Recent methods combine language-model generation with constraint solving,
symbolic decomposition, adaptive code synthesis, and neuro-symbolic validation
~\citep{curtis2025proc3s,petruzzellis2025hvr,chi2025instructflow,
ahn2025reliablecap,lei2025planningdriven,chen2025codesymbolicplanner}.
Other studies connect natural-language specifications with formal planning
through LLM-symbolic pipelines, structured planning languages, task-and-motion
checking, and embodied intermediate reasoning
~\citep{huang2025planningdark,zuo2025planetarium,chen2024autotamp,
zawalski2025embodiedcot,pesjak2026robotplanning}.
These methods show the value of checking proposed actions against explicit
constraints rather than relying on the plausibility of generated text.
However, they generally assume that the goal or required operations are
already specified.
FORGE addresses the complementary setting in which the model must infer
recovery obligations from a faulty assembly and generate a verifier-grounded
program that restores the complete target configuration.

\textbf{Structured action generation.}
Structured action generation maps model decisions to machine-readable actions
for tool use, function calling, and agent interaction.
xLAM, ToolACE, and ToolGen improve action selection, tool retrieval, and
argument generation
~\citep{zhang2025xlam,liu2025toolace,wang2025toolgen}.
Stateful interaction is evaluated by BFCL, ToolSandbox, and ACEBench, while
E-Bench and SciAgentGym extend evaluation to multi-step tool orchestration
and dependency-aware scientific workflows
~\citep{patil2025bfcl,lu2025toolsandbox,chen2025acebench,
zheng2026ebench,shen2026sciagentgym}.
These benchmarks highlight the need to evaluate outcomes across a sequence
of state-changing actions, beyond the correctness of individual tool calls.
Nevertheless, their environments rarely model physical interlocking
constraints or fault-conditioned recovery.
FORGE therefore evaluates complete rework programs whose actions must respect
structural dependencies, avoid unnecessary intervention, and terminate in a
verifier-confirmed target state.

\section{Method}
\label{sec:method}

\subsection{Task Formulation and Framework Overview}
\label{sec:task_formulation}

FORGE studies fault-conditioned executable rework generation for interlocking
assemblies. Given an observed assembly state, a target configuration, and
public structural constraints, the model must determine whether rework is
required, identify the underlying fault, and generate an executable program
that restores the assembly to the target state. Built on vanilla SFT, FORGE
uses Rework Structuring for dependency-aware recovery and Action Grounding for
state-conditioned executable action generation.

We formulate each task as a structured state-to-program generation problem:
\begin{equation}
\begin{aligned}
X &=
\left(
S^{\mathrm{initial}},
S^{\mathrm{actual}},
S^{\mathrm{target}},
\mathcal{G},
\mathcal{A},
P
\right), \\
Y &= (D,\pi),
\qquad
\pi=\bigl((a_t,\theta_t)\bigr)_{t=1}^{T},
\end{aligned}
\label{eq:task_io}
\end{equation}
where $S^{\mathrm{initial}}$ is the initial assembly state,
$S^{\mathrm{actual}}$ is the observed state,
$S^{\mathrm{target}}$ is the target configuration,
$\mathcal{G}$ denotes the public assembly, blocking, contact, and
constraint relations, $\mathcal{A}$ is the available action space, $P$
contains public part information, $D$ denotes the fault diagnosis, and $\pi$
is the executable rework program.

FORGE represents program execution as a symbolic state trajectory:
\begin{equation}
\begin{aligned}
\tau_\pi
&=
\left(
S_0,(a_1,\theta_1),S_1,\ldots,
(a_T,\theta_T),S_T
\right), \\
S_0
&=
S^{\mathrm{actual}},
\qquad
S_t
=
\mathcal{T}
\left(
S_{t-1},a_t,\theta_t;\mathcal{G}
\right),
\quad t=1,\ldots,T,
\end{aligned}
\label{eq:rework_trajectory}
\end{equation}
where $\mathcal{T}$ is the frozen symbolic transition function used for
supervision construction and benchmark evaluation.

A generated program is valid only when its schema is well formed, every
operation is legal in its execution state, and the terminal state satisfies
the target configuration:
\begin{equation}
\operatorname{Valid}(X,Y)
=
\mathbb{I}\!\left[\operatorname{Schema}(Y)\right]
\left(
\prod_{t=1}^{T}
\mathbb{I}\!\left[
(a_t,\theta_t)
\in
\mathcal{A}(S_{t-1};\mathcal{G})
\right]
\right)
\mathbb{I}\!\left[
S_T\equiv_{\mathcal{G}}S^{\mathrm{target}}
\right].
\label{eq:program_validity}
\end{equation}

For an already correct assembly, the safe output is
$D=\textsc{NoFault}$ and $\pi=\varnothing$, which prevents unnecessary
intervention.

\begin{figure}[t]
    \centering
    \includegraphics[width=\linewidth]{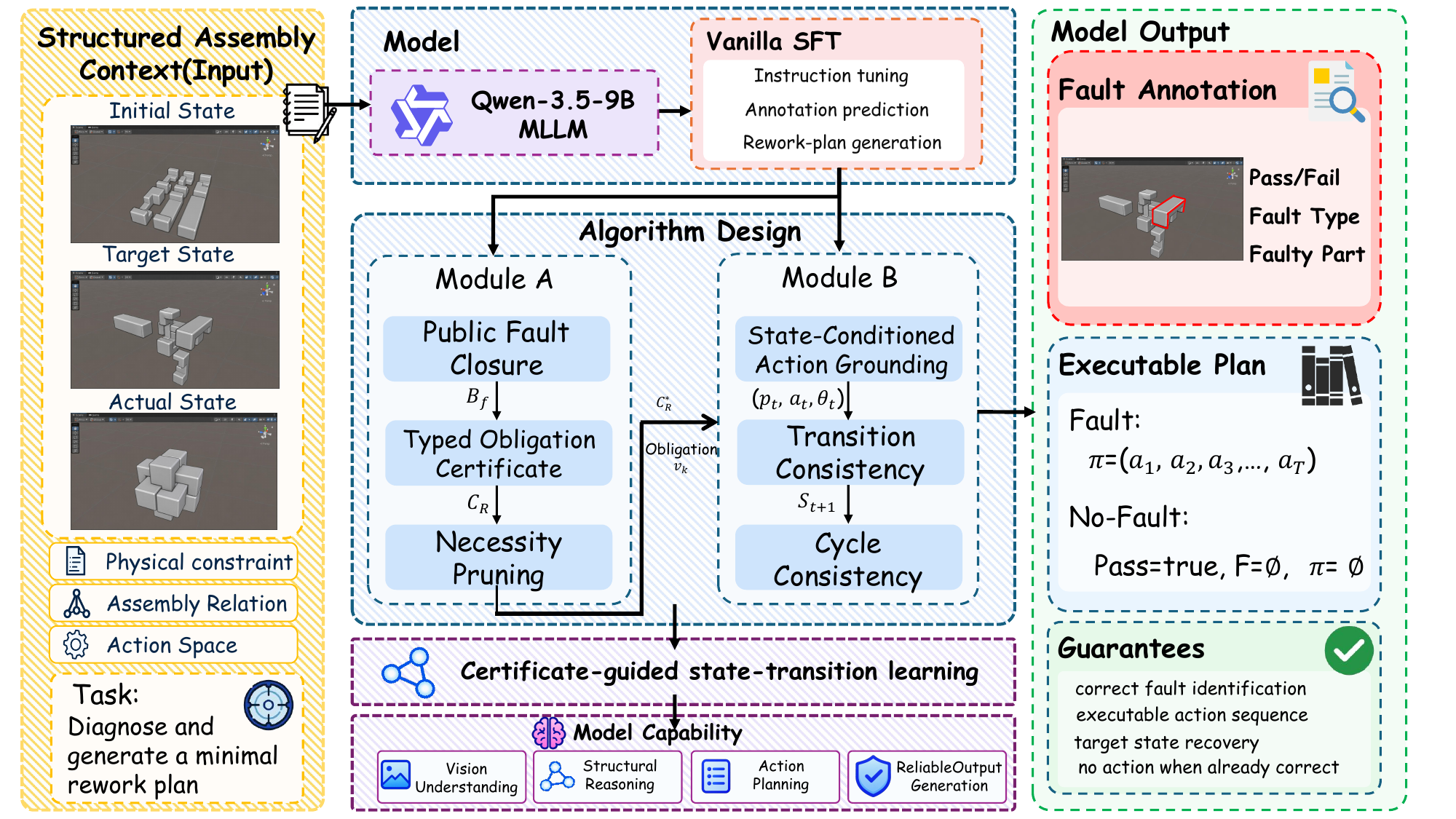}
    \caption{
    Overview of FORGE. Rework Structuring identifies recovery obligations,
    while Action Grounding converts them into executable actions.
    }
    \label{fig:method}
\end{figure}

Vanilla SFT directly learns the diagnosis and rework sequence but does not
explicitly model recovery dependencies or state-conditioned execution.
FORGE therefore augments the SFT backbone with two algorithmic modules:
\textbf{Rework Structuring} and \textbf{Action Grounding}, as illustrated in
Figure~\ref{fig:method}.

\subsection{Fault-Oriented Rework Structuring}
\label{sec:rework_structuring}

Rework Structuring organizes the supervision target before action generation.
It identifies the components and prerequisite relations that are relevant to a
diagnosed fault, but it does not introduce an additional prediction head or
an additional output object.

Given a diagnosed fault set $F$, the relevant component set is expanded using
the public graph relations:
\begin{equation}
\begin{aligned}
D_F^{0}
&=F, \\
D_F^{k+1}
&=
D_F^k
\cup N_b(D_F^k)
\cup N_c(D_F^k)
\cup N_k(D_F^k), \\
\mathcal{D}_F
&=
\min\left\{
D_F^k:
D_F^{k+1}=D_F^k
\right\}.
\end{aligned}
\label{eq:fault_closure}
\end{equation}

The resulting dependency information is used to organize the actions in the
reference \texttt{rework\_plan} and to provide structured supervision for
recovery generation. The intermediate dependency structure is not exposed as
an additional output field; the model generates the final
\texttt{annotation}--\texttt{rework\_plan} object.

\subsection{State-Conditioned Action Grounding}
\label{sec:action_grounding}

Action Grounding conditions each generated action on the symbolic state
produced by preceding operations. The model generates the complete
\texttt{annotation}--\texttt{rework\_plan} object, whose actions are
subsequently evaluated through symbolic replay.

For a current state $S_{t-1}$, the executable action set is
$
\mathcal{A}_t
=
\mathcal{A}
\left(
S_{t-1};\mathcal{G}
\right),
\label{eq:grounded_action_set}
$
and the generated action must satisfy
$
(a_t,\theta_t)\in\mathcal{A}_t.
\label{eq:grounded_action_validity}
$
The training target remains the final diagnosis and rework program:
\begin{equation}
p_{\theta}(D,\pi\mid X)
=
p_{\theta}(D\mid X)
p_{\theta}(\pi\mid X,D).
\label{eq:hierarchical_grounding}
\end{equation}

The decomposition above describes the supervision and execution structure.
All fields are generated by the same autoregressive model and serialized into
the final \texttt{annotation}--\texttt{rework\_plan} schema.

\subsection{Training Objective and Inference Protocol}
\label{sec:training_inference}
FORGE uses Rework Structuring to organize fault-dependent recovery sequences
and Action Grounding to condition each action on its preceding symbolic state.
The resulting diagnosis and rework program are serialized into a single
autoregressive training target without separate prediction heads or auxiliary
losses:
\begin{equation}
\begin{aligned}
Y
&=
\operatorname{Serialize}(D,\pi) \\
&=
\operatorname{BOS}
\mathbin{\Vert}
\operatorname{Encode}(D)
\mathbin{\Vert}
\operatorname{SEP}
\mathbin{\Vert}
\operatorname{Encode}(\pi)
\mathbin{\Vert}
\operatorname{EOS} \\
&=
(y_1,\ldots,y_{|Y|}),
\end{aligned}
\label{eq:serialized_target}
\end{equation}
We train the model with teacher-forced supervised fine-tuning:
\begin{equation}
\mathcal{L}_{\mathrm{SFT}}(\theta)
=
-\sum_{(X,Y)\in\mathcal{D}}
\sum_{t=1}^{|Y|}
\log p_{\theta}\!\left(y_t\mid X,y_{<t}\right).
\label{eq:unified_sft}
\end{equation}

During inference, the model generates the same final schema using deterministic
autoregressive decoding. The generated sequence is then parsed and replayed
under the symbolic transition rules:
\begin{equation}
\begin{aligned}
\hat{Y}&=\operatorname{Decode}_{\mathrm{greedy}}(X),\\
(\hat{D},\hat{\pi})&=\operatorname{Parse}(\hat{Y}).
\end{aligned}
\label{eq:deterministic_decoding}
\end{equation}

The symbolic transition engine then checks schema validity, action legality,
state transitions, and target-state restoration.
\section{Dataset}
\label{sec:dataset}

We introduce \textsc{FORGE-Bench}, a benchmark for fault-conditioned
executable rework generation in interlocking assemblies. Unlike conventional
assembly datasets that focus on forward construction, assembly-sequence
prediction, or instruction generation, \textsc{FORGE-Bench} evaluates whether
a model can diagnose faulty assembly sEach task provides a public assembly
representation comprising the initial state, observed state, target state,
structural relations, constraints, action space, and part catalog.Given this information,
the model must determine whether rework is required and, when a fault is
present, generate a program that restores the target configuration.

\subsection{Dataset Construction}
\label{sec:dataset_construction}

\begin{figure}[ht]
    \centering
    \includegraphics[
        width=\linewidth
    ]{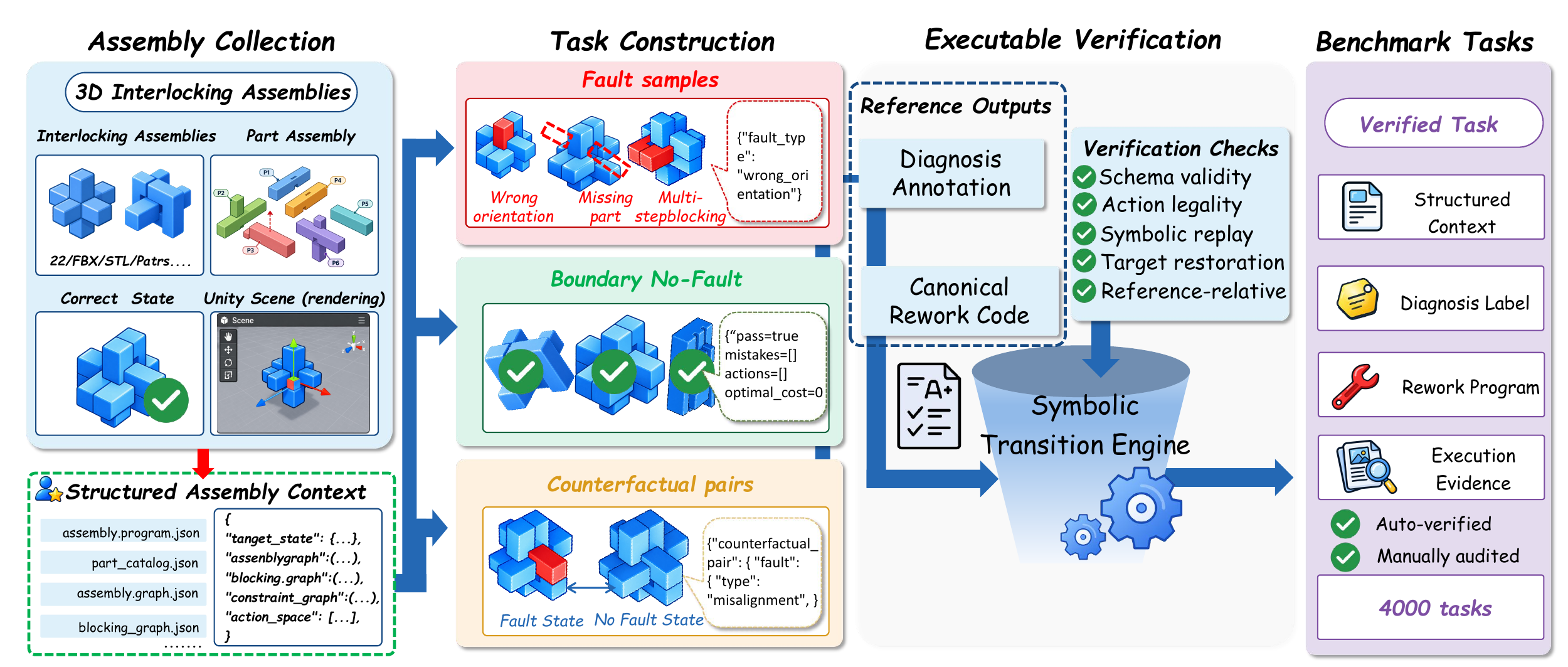}
    \caption{
    Construction pipeline of \textsc{FORGE-Bench}. Executable assembly
    structures are converted into public task representations, augmented with
    controlled fault and no-fault states, paired with reference rework
    programs, and filtered through symbolic execution and target-state
    restoration.
    }
    \label{fig:dataset_construction}
\end{figure}

As shown in Figure~\ref{fig:dataset_construction},
\textsc{FORGE-Bench} is constructed as a closed
execution-and-verification loop rather than a static collection of assembly
states. The pipeline converts executable assembly structures into
fault-conditioned rework tasks and retains only instances whose reference
programs satisfy structural, action-level, and target-restoration constraints.

\noindent\textbf{Assembly collection and executable representation.}
We collect 3D interlocking assembly structures and instantiate their valid
configurations as executable Unity scenes. Each assembly is represented by a
structured context containing its part catalog, assembly relations, blocking
dependencies, physical constraints, and available action interfaces. This
context defines both the target configuration and the conditions under which
individual operations can be executed.

\noindent\textbf{Fault-conditioned task construction.}
From each assembly context, we generate faulty states, boundary no-fault states,
and matched fault/no-fault pairs. Faults include incorrect orientations,
missing components, misplacements, and errors hidden behind multi-step blocking
dependencies. Boundary no-fault states remain within the target tolerance and
therefore require an empty rework program.

\noindent\textbf{Reference program generation and verification.}
Each task is paired with a diagnosis annotation and a verified
sequence-minimal rework program. For a faulty state, the program resolves the
relevant blocking dependencies, corrects the fault, and restores temporarily
affected components. A frozen symbolic transition engine replays every action
and checks schema validity, action legality, state-transition consistency, and
target-state restoration.

\subsection{Dataset Analysis}
\label{sec:dataset_analysis}

{
\setlength{\intextsep}{2pt}
\begin{figure}[H]
    \centering

    \begin{minipage}[t]{0.54\linewidth}
        \vspace{0pt}
        \small
Figure~\ref{fig:dataset_composition} and
Table~\ref{tab:dataset_statistics} summarize the composition and
scale of \textsc{FORGE-Bench}. \textsc{FORGE-Bench} contains 4{,}000 tasks
across 26 interlocking assemblies, with a 3{,}000/1{,}000 train-test split
and five held-out assemblies for evaluating generalization to unseen
structures. The benchmark includes 1{,}115 no-fault cases (27.9\%), 2{,}353
single-fault cases (58.8\%), and 532 multi-fault cases (13.3\%), combining
routine diagnosis with dependency-coupled recovery at different fault scopes
and structural complexities. Counterfactual metadata contains 160 matched
fault/no-fault pairs, or 320 states, providing paired examples of required
intervention and valid no-intervention cases across diverse structural
configurations and recovery scenarios.
    \end{minipage}
    \hfill
    \begin{minipage}[t]{0.42\linewidth}
        \centering
        \vspace{0pt}

        \includegraphics[
            width=\linewidth,
            height=4.6cm,
            keepaspectratio
        ]{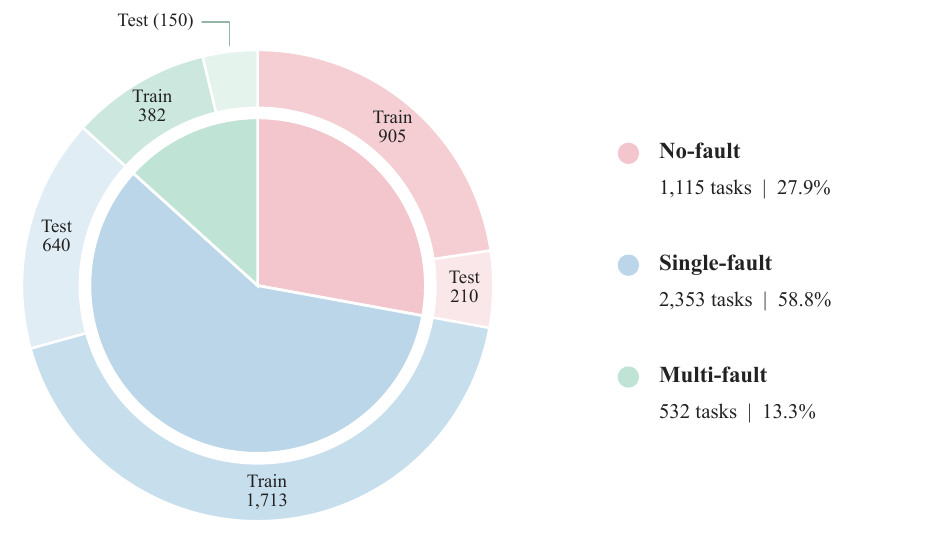}

        \vspace{2mm}

        \caption{Composition of \textsc{FORGE-Bench}.}
        \label{fig:dataset_composition}
    \end{minipage}
\end{figure}
}

\vspace{2mm}

\begin{table}[H]
    \small
    \centering
    \caption{Key statistics of \textsc{FORGE-Bench}.}
    \label{tab:dataset_statistics}
    \renewcommand{\arraystretch}{1.05}
    \setlength{\tabcolsep}{3.2mm}
    \begin{tabular}{lrrlrr}
        \toprule
        \multicolumn{3}{c}{\textbf{Dataset statistics}} &
        \multicolumn{3}{c}{\textbf{Rework statistics}} \\
        \cmidrule(lr){1-3}
        \cmidrule(lr){4-6}
        Statistic & Count & Share &
        Statistic & Count & Share \\
        \midrule
        Total tasks
        & 4{,}000
        & 100.0\%
        &
        No-fault cases
        & 1{,}115
        & 27.9\% \\

        Train / Test
        & 3{,}000 / 1{,}000
        & 75.0 / 25.0\%
        &
        Single-fault cases
        & 2{,}353
        & 58.8\% \\

        Assembly families
        & 26
        & --
        &
        Multi-fault cases
        & 532
        & 13.3\% \\

        Train assemblies
        & 21
        & --
        &
        Counterfactual pairs
        & 160
        & -- \\

        Test assemblies
        & 5
        & --
        &
        Fault/no-fault instances
        & 320
        & -- \\

        \bottomrule
    \end{tabular}
\end{table}

\section{Experiments}
\label{sec:experiments}

\subsection{Experimental Setup}
\label{sec:experimental_setup}

We build \textsc{FORGE} on Qwen3.5-9B-Base~\citep{qwen2025qwen35} and apply LoRA
with rank 8, scaling factor 32, and dropout 0.05 to all linear modules.
Training uses bfloat16 precision, a per-device batch size of 1, and gradient
accumulation over 4 steps. Vanilla SFT provides the training backbone, while
Rework Structuring and Action Grounding support dependency-aware program
generation.
Inference uses deterministic decoding with temperature 0 and a maximum output
length of 4{,}096 tokens, with thinking disabled.

We compare \textsc{FORGE} with open-source models, including
Qwen3.5-9B Base~\citep{qwen2025qwen35},
Gemma-3-12B-IT~\citep{team2025gemma3},
GLM-4-9B-0414-Chat~\citep{glm2025glm4},
InternLM3-8B-Instruct~\citep{cai2024internlm3},
Ministral-3-8B-Instruct~\citep{ministral2025}, and
Llama-3.1-8B-Instruct~\citep{dubey2024llama3};
closed-source models, including
GPT-5.4 Mini~\citep{OpenAI2026GPT5_4},
Claude Sonnet 4.6~\citep{Anthropic2026ClaudeSonnet4_6},
Qwen3.6-Plus~\citep{qwen3.6-35b-a3b}, and
Gemini 3.5 Flash~\citep{GoogleDeepMind2026Gemini3_5Flash};
and code-oriented methods, including
AceCoder-7B~\citep{zeng2025acecoder},
OpenCodeReasoning-Nemotron-7B~\citep{ahmad2025opencodereasoning}, and
Seed-Coder-8B-Reasoning~\citep{zhang2025seedcoder}.
All models receive the same public task representation and produce the shared
\texttt{annotation}--\texttt{rework\_plan} schema. Generated programs are
evaluated with the same symbolic transition engine under the benchmark
protocol, while unadapted Qwen3.5-9B serves as the backbone baseline. We
evaluate all methods on
FORGE-Bench using six execution-oriented metrics, their unweighted aggregate
Overall-Core, and a separate LLM-as-Judge score. Metric definitions are
provided in Appendix~\ref{app:metrics}.
\subsection{Main Results}
\label{sec:main_results}
\definecolor{bestcolor}{RGB}{217,232,252}
\definecolor{secondcolor}{RGB}{221,242,214}

\begin{table*}[!htbp]
\centering
\footnotesize
\caption{
Main results on \textsc{FORGE-Bench}. Models are grouped into open-source
models, closed-source models, prior methods, and our method. The best results
are highlighted in bold, and the second-best results are underlined.
}
\label{tab:main_results}

\setlength{\tabcolsep}{2.5pt}
\renewcommand{\arraystretch}{0.95}

\resizebox{\textwidth}{!}{
\begin{tabular}{l|cccccc|c|c}
\toprule

\multirow{2}{*}{\textbf{Model}}
& \multicolumn{6}{c|}{\textbf{Core Metrics}}
& \multirow{2}{*}{\textbf{Overall-Core}$\uparrow$}
& \textbf{LLM-as-Judge}
\\

\cmidrule(lr){2-7}

& \textbf{SSE}$\uparrow$
& \textbf{SAPF}$\uparrow$
& \textbf{VMRE}$\uparrow$
& \textbf{DGTR}$\uparrow$
& \textbf{MVE}$\uparrow$
& \textbf{HRQ}$\uparrow$
&
& \textbf{Score}$\uparrow$
\\

\midrule

\multicolumn{9}{c}{\emph{Open-source Models}}
\\
\midrule

Qwen3.5-9B Base~\citep{qwen2025qwen35}
& 20.00 & 11.04 & 22.50 & 25.83 & 22.50 & 20.00
& 20.31 & 2.55
\\

Gemma-3-12B-IT~\citep{team2025gemma3}
& 6.21 & 50.00 & 2.50 & 5.83 & 2.50 & \textemdash
& 11.17 & 1.91
\\

GLM-4-9B-0414-Chat~\citep{glm2025glm4}
& 8.42 & 38.75 & 3.33 & 3.33 & 3.33 & \textemdash
& 9.53 & 1.87
\\

InternLM3-8B-Instruct~\citep{cai2024internlm3}
& 16.00 & 22.22 & 20.00 & 20.00 & 20.00 & 20.00
& 19.70 & 1.79
\\

Ministral-3-8B-Instruct~\citep{ministral2025}
& 18.21 & 27.78 & 17.50 & 20.83 & 17.50 & 20.00
& 20.30 & 2.42
\\

Llama-3.1-8B-Instruct~\citep{dubey2024llama3}
& 5.50 & 57.36 & \textemdash & \textemdash & \textemdash & \textemdash
& 10.48 & 1.40
\\

\midrule
\multicolumn{9}{c}{\emph{Closed-source Models}}
\\
\midrule

GPT-5.4 Mini~\citep{OpenAI2026GPT5_4}
& 25.50 & 37.15 & 25.16 & 25.74 & 25.16 & 30.63
& 28.22 & 2.94
\\

Claude Sonnet 4.6~\citep{Anthropic2026ClaudeSonnet4_6}
& 31.35 & 41.11 & 27.87 & 36.19 & 25.22 & 23.43
& 30.86 & 4.00
\\

Qwen3.6-Plus~\citep{qwen3.6-35b-a3b}
& 22.55 & 47.22 & 18.77 & 19.44 & 18.19 & 14.37
& 23.43 & 2.72
\\

Gemini 3.5 Flash~\citep{GoogleDeepMind2026Gemini3_5Flash}
& \cellcolor{secondcolor}\underline{62.93}
& \cellcolor{bestcolor}\textbf{69.10}
& \cellcolor{secondcolor}\underline{28.12}
& \cellcolor{secondcolor}\underline{45.49}
& \cellcolor{secondcolor}\underline{26.61}
& 46.39
& \cellcolor{secondcolor}\underline{46.44}
& \cellcolor{secondcolor}\underline{4.26}
\\

\midrule
\multicolumn{9}{c}{\emph{Prior Methods}}
\\
\midrule

AceCoder-7B~\citep{zeng2025acecoder}
& 28.89 & 8.33 & 26.14 & 29.27 & 26.14
& \cellcolor{secondcolor}\underline{52.58}
& 28.56 & 3.63
\\

OpenCodeReasoning-Nemotron-7B~
\citep{ahmad2025opencodereasoning}
& 15.25 & 16.67 & 15.21 & 15.21 & 15.21 & 9.55
& 14.51 & 2.34
\\

Seed-Coder-8B-Reasoning~\citep{zhang2025seedcoder}
& 8.57 & 6.67 & 9.83 & 11.67 & 8.00 & \textemdash
& 7.46 & 2.61
\\

\midrule
\textbf{FORGE (Ours)}
& \cellcolor{bestcolor}\textbf{71.54}
& \cellcolor{secondcolor}\underline{61.11}
& \cellcolor{bestcolor}\textbf{72.60}
& \cellcolor{bestcolor}\textbf{75.93}
& \cellcolor{bestcolor}\textbf{72.13}
& \cellcolor{bestcolor}\textbf{83.17}
& \cellcolor{bestcolor}\textbf{72.75}
& \cellcolor{bestcolor}\textbf{4.37}
\\

\bottomrule
\end{tabular}
}
\end{table*}

Table~\ref{tab:main_results} shows that \textsc{FORGE} achieves the highest
Overall-Core score of 72.75 among all compared methods. It obtains the best performance on five of the six core
metrics, including SSE, VMRE, DGTR, MVE, and HRQ. These gains indicate that
the proposed framework improves both recovery structure and executable target
restoration.

The independent LLM-as-Judge evaluation follows the same trend, with
\textsc{FORGE} achieving the highest score of 4.37. The largest improvements
appear on execution-oriented metrics, suggesting that \textsc{FORGE} more
effectively converts structured repair intent into executable recovery
programs.
\subsection{Ablation Study}
\label{sec:ablation}

\begin{table*}[!htbp]
\centering
\footnotesize
\caption{
Ablation study of FORGE components. Each component-removal variant retains
the other component. The best results are highlighted in bold with a blue
background.
}
\label{tab:ablation_results}

\setlength{\tabcolsep}{2.5pt}
\renewcommand{\arraystretch}{0.95}

\resizebox{\textwidth}{!}{
\begin{tabular}{l|cccccc|c|c}
\toprule

\multirow{2}{*}{\textbf{Method}}
& \multicolumn{6}{c|}{\textbf{Core Metrics}}
& \multirow{2}{*}{\textbf{Overall-Core}$\uparrow$}
& \textbf{LLM-as-Judge}
\\

\cmidrule(lr){2-7}

& \textbf{SSE}$\uparrow$
& \textbf{SAPF}$\uparrow$
& \textbf{VMRE}$\uparrow$
& \textbf{DGTR}$\uparrow$
& \textbf{MVE}$\uparrow$
& \textbf{HRQ}$\uparrow$
&
& \textbf{Score}$\uparrow$
\\

\midrule

Vanilla SFT
& 69.23
& 61.11
& 63.23
& 71.56
& 62.60
& 77.44
& 67.53
& 4.03
\\

\textit{w/o Action Grounding}
& 62.38
& 41.67
& 62.19
& 68.19
& 60.10
& 73.60
& 61.35
& 4.12
\\

\textit{w/o Rework Structuring}
& 65.43
& \cellcolor{bestcolor}\textbf{66.67}
& 64.05
& 67.38
& 63.11
& 80.67
& 67.89
& 4.30
\\

\midrule

\textbf{FORGE (Ours)}
& \cellcolor{bestcolor}\textbf{71.54}
& 61.11
& \cellcolor{bestcolor}\textbf{72.60}
& \cellcolor{bestcolor}\textbf{75.93}
& \cellcolor{bestcolor}\textbf{72.13}
& \cellcolor{bestcolor}\textbf{83.17}
& \cellcolor{bestcolor}\textbf{72.75}
& \cellcolor{bestcolor}\textbf{4.37}
\\

\bottomrule
\end{tabular}
}
\end{table*}

Table~\ref{tab:ablation_results} compares Vanilla SFT, the two
single-component variants, and full \textsc{FORGE}.
Vanilla SFT achieves an Overall-Core score of 67.53, while full
\textsc{FORGE} reaches 72.75.
Removing Action Grounding, while retaining Rework Structuring, reduces
Overall-Core to 61.35, with SAPF and HRQ falling to 41.67 and 73.60,
respectively.
Removing Rework Structuring, while retaining Action Grounding, yields an
Overall-Core score of 67.89; DGTR and MVE decrease from 75.93 and 72.13
for the full model to 67.38 and 63.11, respectively.
This variant nevertheless achieves the highest SAPF of 66.67, indicating
that parameter fidelity alone does not ensure successful target restoration.
Full \textsc{FORGE} achieves the highest SSE, VMRE, DGTR, MVE, HRQ,
Overall-Core, and LLM-as-Judge scores among the compared configurations.
These results support the joint use of recovery structuring and action
grounding, while showing that their benefits are not uniform across
all metrics.

\subsection{Case Study and Error Analysis}
\label{sec:case_study}

\begin{figure}[ht]
    \centering
    \includegraphics[
        width=\linewidth
    ]{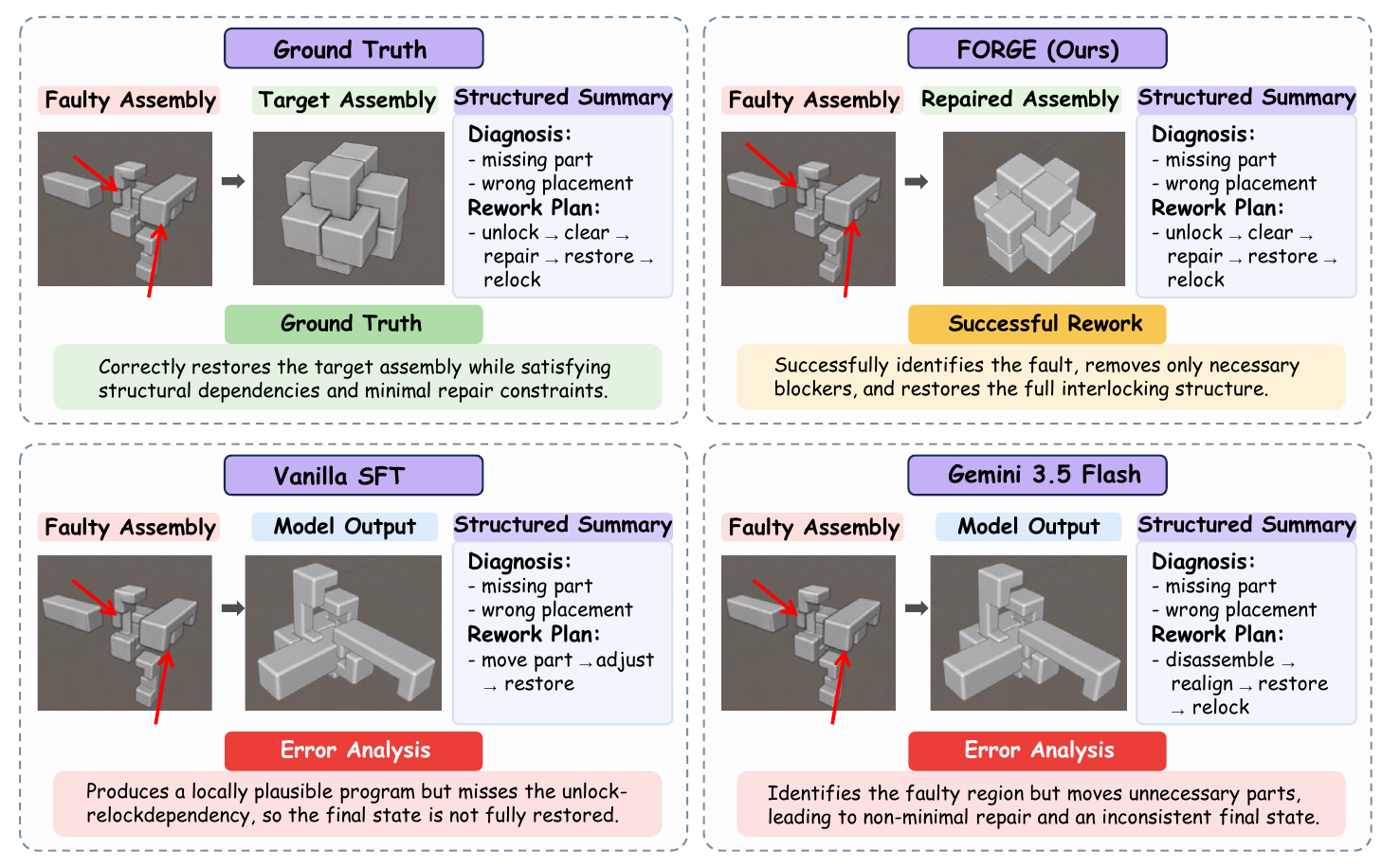}
    \caption{
    Qualitative comparison of executable rework generation.
    \textsc{FORGE} preserves recovery dependencies and restores the target
    assembly, while baseline predictions exhibit missing dependency handling
    or unnecessary intervention.
    }
    \label{fig:case_study}
\end{figure}

Figure~\ref{fig:case_study} presents a representative interlocking assembly
rework case. The example highlights the key challenge of executable rework
generation: models must identify faulty components and preserve recovery
dependencies during restoration. \textsc{FORGE} generates a dependency-aware
repair procedure, while baseline models exhibit missing dependency handling
and unnecessary intervention. This comparison supports the quantitative
finding that structured recovery reasoning and state-conditioned action
grounding improve executable assembly restoration.

\vspace{-2mm}

\subsection{Consistency with Human Judgments}
\label{sec:consistency_human}

\begin{figure}[H]
\centering

\begin{minipage}[t]{0.52\columnwidth}
\vspace{0pt}
Figure~\ref{fig:human_consistency} shows a clear positive alignment between
automatic evaluation and expert judgment across the evaluated models. Most
baseline systems follow the rising trend from the lower-left to the
upper-middle region, with weaker verified execution and incomplete target
restoration receiving lower expert ratings. \textsc{FORGE} is located in the
upper-right region, indicating that its benchmark improvements are reflected
in expert assessment. This ordering suggests that the metrics capture
structural recovery, executable action sequences, and complete target-state
restoration across the evaluated assemblies and fault types, rather than
surface-level plausibility alone.
\end{minipage}
\hfill
\begin{minipage}[t]{0.43\columnwidth}
\vspace{0pt}

\centering
\includegraphics[width=\linewidth]{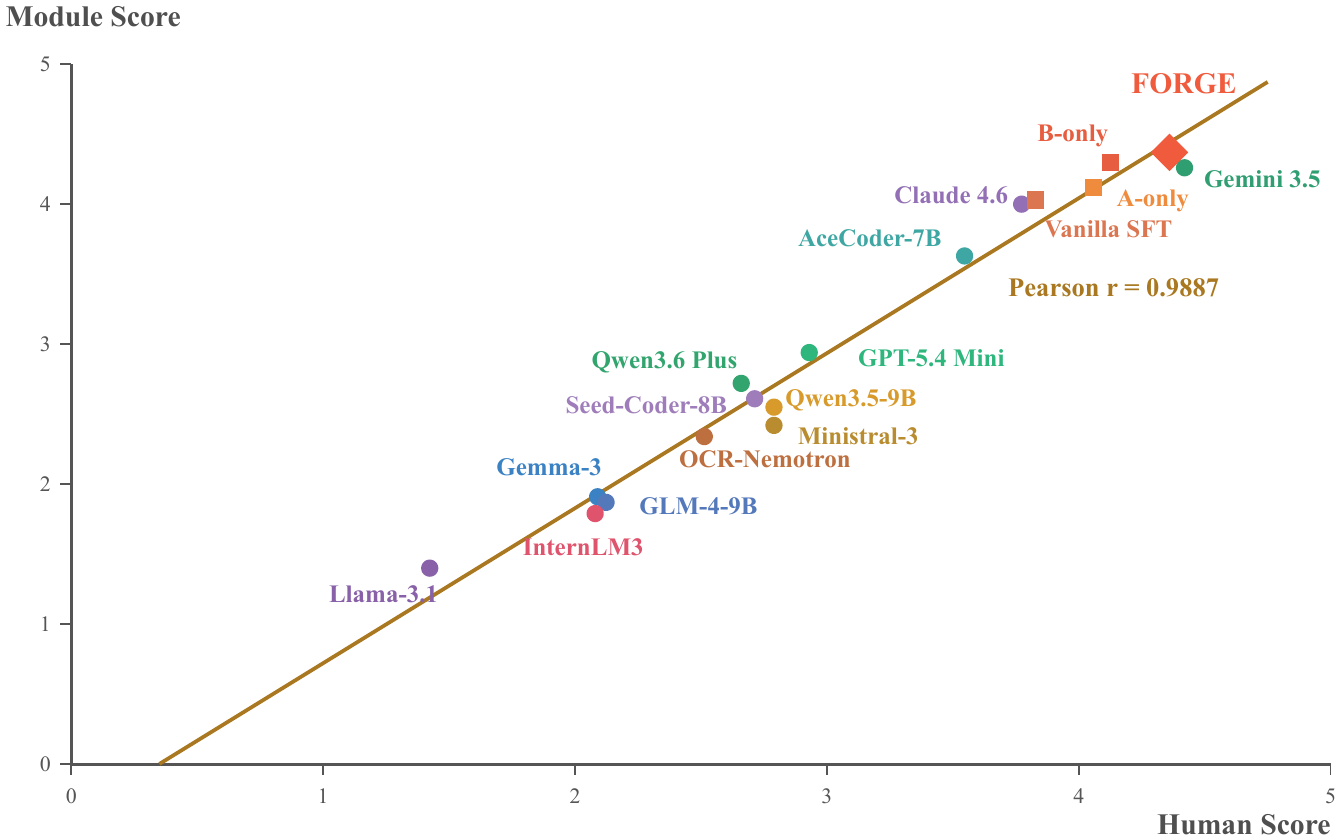}

\caption{
Consistency between automatic evaluation and human judgment.
}
\label{fig:human_consistency}

\end{minipage}

\end{figure}

\section{Conclusion and Limitations}
\label{sec:conclusion}

We introduce \emph{fault-conditioned executable rework generation}, where assembly repair shifts from identifying an isolated fault to producing a complete recovery program that respects structural dependencies and restores the target configuration. \emph{FORGE-Bench} makes this capability measurable through structured interlocking assembly tasks, controlled fault and no-fault conditions, and verifier-backed execution, while \emph{FORGE} addresses it through rework structuring and action grounding. Experiments reveal a clear \emph{dependency-execution gap}: existing models can often identify the faulty component or propose a plausible local action, but fail to resolve blocking dependencies, preserve displaced components, or restore the complete assembly state. \emph{FORGE} narrows this gap by coupling fault understanding with executable program generation and target-state verification. This work focuses on symbolic interlocking assembly environments with discrete action interfaces and a finite set of assembly structures, so extending the same principle to richer assembly families, continuous physical execution, uncertain perception, and interactive recovery with execution feedback will require additional action models, environment adapters, and validity criteria, and remains a natural direction for future work.

\clearpage
\section*{AI Use Statement}

Generative AI assistance was used for literature processing, dataset
construction, and auxiliary LLM-as-Judge evaluation.
All generated records, annotations, and evaluation outputs were reviewed under
the documented audit procedure.
The authors are responsible for the final dataset, experimental results,
analyses, and claims.

\section*{Reproducibility Statement}

\hyperref[app:task]{Appendix~\ref*{app:task}} defines the FORGE-Bench
task contract. \hyperref[app:dataset]{Appendix~\ref*{app:dataset}}
describes dataset construction and split protocols;
\hyperref[app:training]{Appendix~\ref*{app:training}} gives the training
and inference configuration; and
\hyperref[app:metrics]{Appendix~\ref*{app:metrics}} defines the evaluation
metrics. All reported results use the fixed split and evaluation protocol
specified in \hyperref[app:dataset]{Appendix~\ref*{app:dataset}}.

\bibliography{iclr2027_conference}
\bibliographystyle{iclr2027_conference}

\clearpage
\appendix

\section{Task Definition and Output Schema}
\label{app:task}

This appendix specifies the public task contract used by FORGE-Bench. TThe benchmark provides an initial assembly state, an observed assembly state,
a target state, structural relations, an action library, and a public part
catalog.The model must decide
whether rework is required, diagnose the fault when present, and produce a
state-valid rework program.

\subsection{Public Input Contract}
\label{app:task:input}

Each task is represented as
\[
X=(S^{\mathrm{initial}},S^{\mathrm{actual}},S^{\mathrm{target}},
\mathcal{G},\mathcal{A},P),
\]
where $S^{\mathrm{initial}}$ is the initial assembly state,
$S^{\mathrm{actual}}$ is the observed state, $S^{\mathrm{target}}$ is
the target configuration,$\mathcal{G}$ contains assembly, blocking, contact,
and constraint relations, $\mathcal{A}$ is the action space, and $P$ is the
public part catalog. The benchmark package contains the following files:

\begin{table}[ht]
\centering
\small
\caption{Public files in a FORGE-Bench task.}
\label{tab:appendix_public_files}
\begin{tabular}{p{0.30\linewidth}p{0.56\linewidth}}
\toprule
\textbf{File} & \textbf{Role} \\
\midrule
\texttt{initial\_state.json} & Initial assembly state. \\
\texttt{target\_state.json} & Target assembly configuration. \\
\texttt{actual\_state.json} & Observed state supplied to the model. \\
\texttt{assembly\_graph.json} & Assembly relations between parts. \\
\texttt{blocking\_graph.json} & Blocking and access dependencies. \\
\texttt{contact\_graph.json} & Contact or adjacency relations. \\
\texttt{action\_space.json} & Available actions and their interfaces. \\
\texttt{part\_catalog.json} & Part identifiers and public metadata. \\
\bottomrule
\end{tabular}
\end{table}

\subsection{Output Contract}
\label{app:task:output}

The structured target contains a diagnosis and a rework plan. The executable
program is written as $\pi=((a_t,\theta_t))_{t=1}^{T}$, where $a_t$ is an
action type and $\theta_t$ contains its parameters.

A compressed real task example from the repository has the following form:

\begin{verbatim}
{
  "task_id": "bayonet_structure_train_000000",
  "assembly_id": "bayonet_structure",
  "input_contract": [
    "initial_state.json", "target_state.json", "actual_state.json",
    "assembly_graph.json", "blocking_graph.json", "contact_graph.json",
    "action_space.json", "part_catalog.json"
  ],
  "actual_state": {
    "BayonetStructure_Part01_Lock": {"present": true},
    "BayonetStructure_Part02_Pin": {"present": false}
  }
}
\end{verbatim}

The model returns one JSON object with the top-level fields
\texttt{annotation} and \texttt{rework\_plan}. For the example above, the
corresponding output is:

\begin{verbatim}
{
  "annotation": {
    "sample_id": "bayonet_structure_train_000000",
    "assembly_id": "bayonet_structure",
    "pass": false,
    "mistakes": [{
      "type": "missing_part",
      "part_id": "BayonetStructure_Part02_Pin",
      "visible": true
    }]
  },
  "rework_plan": {
    "objective": "restore_to_target_state",
    "actions": [
      {
        "action": "insert",
        "part_id": "BayonetStructure_Part02_Pin"
      }
    ]
  }
}
\end{verbatim}

For a no-fault sample,
\texttt{annotation.pass=true},
\texttt{annotation.mistakes=[]}, and
\texttt{rework\_plan.actions=[]}.

\subsection{Action Library and Preconditions}
\label{app:task:actions}

The recorded action types are \texttt{remove}, \texttt{insert},
\texttt{reset\_pose}, \texttt{rotate}, \texttt{rotate\_to\_unlock},
\texttt{rotate\_translate\_to\_unlock}, \texttt{rotate\_to\_lock},
\texttt{translate\_rotate\_to\_lock}, and \texttt{replace}.

\begin{table}[ht]
\centering
\scriptsize
\setlength{\tabcolsep}{2pt}
\caption{Recorded action interfaces and symbolic preconditions.}
\label{tab:appendix_actions}
\begin{tabular}{@{}p{0.43\linewidth}p{0.49\linewidth}@{}}
\toprule
\textbf{Action} & \textbf{Recorded precondition} \\
\midrule
\texttt{remove(part\_id)} & Part is present and not blocked for removal. \\
\texttt{insert(part\_id)} & Part is absent, target pose is free, and insertion path is collision-free. \\
\texttt{reset\_pose(part\_id)} & Part is present and reset path is collision-free. \\
\begin{tabular}[t]{@{}l@{}}\texttt{rotate(part\_id, axis,}\\[-1pt]\texttt{angle\_deg)}\end{tabular}
& Part is present and rotation path is collision-free. \\
\begin{tabular}[t]{@{}l@{}}\texttt{rotate\_to\_unlock(part\_id,}\\[-1pt]\texttt{axis, angle\_deg)}\end{tabular}
& Part is present and the rotation path is collision-free. \\
\begin{tabular}[t]{@{}l@{}}\texttt{rotate\_translate\_to\_unlock}\\[-1pt]\texttt{(part\_id, after\_pose)}\end{tabular}
& Part is present and the coupled unlock path is collision-free. \\
\begin{tabular}[t]{@{}l@{}}\texttt{rotate\_to\_lock(part\_id,}\\[-1pt]\texttt{axis, angle\_deg)}\end{tabular}
& Part is present and the assembly has been restored. \\
\begin{tabular}[t]{@{}l@{}}\texttt{translate\_rotate\_to\_lock}\\[-1pt]\texttt{(part\_id, after\_pose)}\end{tabular}
& Part is present and the assembly has been restored. \\
\begin{tabular}[t]{@{}l@{}}\texttt{replace(wrong\_part\_id,}\\[-1pt]\texttt{target\_part\_id)}\end{tabular}
& Wrong part is present, target part is absent, and the replacement path is collision-free. \\
\bottomrule
\end{tabular}
\end{table}

An action is legal only in the symbolic state produced by all preceding
actions:
\begin{equation}
\operatorname{Legal}(a_t,\theta_t\mid S_{t-1},\mathcal{G})
=
\mathbb{I}\!\left[
(a_t,\theta_t)\in
\mathcal{A}(S_{t-1};\mathcal{G})
\right].
\label{eq:appendix_action_legality}
\end{equation}
The symbolic transition is
\begin{equation}
S_t=\mathcal{T}(S_{t-1},a_t,\theta_t;\mathcal{G}).
\label{eq:appendix_state_transition}
\end{equation}

\section{FORGE-Bench Construction Details}
\label{app:dataset}

\subsection{Assembly Collection and Representation}
\label{app:dataset:assembly}

The benchmark uses Luban lock interlocking assemblies. The repository contains
parametric assembly descriptions, Unity scenes, assembly graphs, blocking
graphs, and part catalogs. The reported benchmark contains 26 assemblies,
with 21 training assemblies and 5 held-out test assemblies. The assembly-local
coordinate convention is $y$-up; lengths are expressed in Unity scene units.
All headline tables and claims in this manuscript use the final 4,000-task
protocol with a 3,000/1,000 train/test split. The split contains 21 training
assemblies and 5 held-out test assemblies, with no assembly identity shared
across the two partitions.

\subsection{Fault Taxonomy and Task Composition}
\label{app:dataset:faults}

The benchmark covers no-fault, single-fault, and multi-fault conditions. Faults
include wrong orientation, missing parts, misplacement, blocking dependencies,
wrong-part substitutions, under-insertion, and interference locks. Multi-fault
instances contain two or more concurrent fault conditions, while no-fault
instances require an empty rework plan. Counterfactual pairs provide matched fault and no-fault states as auxiliary benchmark metadata.

\begin{table}[ht]
\centering
\small
\caption{FORGE-Bench composition under the 4,000-task protocol.}
\label{tab:appendix_dataset_statistics}
\begin{tabular}{lrr}
\toprule
\textbf{Statistic} & \textbf{Count} & \textbf{Share / note} \\
\midrule
Total tasks & 4{,}000 & 100.0\% \\
Train / test tasks & 3{,}000 / 1{,}000 & 75.0 / 25.0\% \\
Assembly families & 26 & 21 train, 5 held-out \\
No-fault cases & 1{,}115 & 27.9\% \\
Single-fault cases & 2{,}353 & 58.8\% \\
Multi-fault cases & 532 & 13.3\% \\
Counterfactual pairs & 160 & Auxiliary evaluation metadata \\
Fault/no-fault instances & 320 & Auxiliary evaluation metadata \\
\bottomrule
\end{tabular}
\end{table}

\subsection{Reference Program Generation}
\label{app:dataset:reference}

Reference programs are constructed from the assembly program, the validated
disassembly sequence, the blocking graph, and the target state. Each reference
program follows the fixed discrete action grammar and resolves the relevant
blocking dependencies before restoring the affected components. The number of
executable actions in this verified reference sequence defines the reference
cost $c_i^{*}$ used by the rework metrics.

We refer to this construction as \emph{reference-sequence minimality}: the
reference uses the fewest executable actions required by the prescribed
disassembly--repair--restoration sequence under the benchmark's symbolic
transition rules. A valid assembly is assigned an empty reference program.

\subsection{Symbolic Verification and Filtering}
\label{app:dataset:verification}

The frozen transition engine replays every action and checks:

\begin{enumerate}[leftmargin=2em]
    \item output-contract validity;
    \item action preconditions and action legality;
    \item parameter validity under the declared tolerances;
    \item consistency of every symbolic state transition;
    \item target-state restoration; and
    \item empty-program compliance for no-fault cases and reference-relative cost where applicable.
\end{enumerate}

The verifier records contract validity, symbolic executability, target-state
restoration, and structured failure reasons for each prediction.

\begin{verbatim}
Input: observed state S0, target state S*, context G, program pi
1. Parse pi and reject malformed output.
2. For each action (a_t, theta_t) in pi:
3.     Check Legal(a_t, theta_t | S_{t-1}, G).
4.     If illegal, reject and record the failure reason.
5.     Compute S_t = T(S_{t-1}, a_t, theta_t; G).
6. Check diagnosis, parameter validity, and target equivalence.
7. Check no-fault abstention and reference-relative cost where required.
8. Return pass/fail, final state, and structured failure reasons.
\end{verbatim}

\subsection{Split Protocol}
\label{app:dataset:split}

The split is assembly-disjoint: train and test use different assembly
identities. Assembly IDs, the sample-generation configuration, and the split
manifest are fixed for the reported experiments.

The training and formal inference records use random seed 42. The split
manifest is fixed for all reported experiments.

\section{Training and Inference Details}
\label{app:training}

\subsection{Model and Optimization}
\label{app:training:model}

FORGE is initialized from Qwen3.5-9B at \texttt{/home/models/Qwen3.5-9B}.
LoRA is applied to all linear modules with rank 8, scaling factor 32, and
dropout 0.05. Training uses bfloat16 precision, per-device batch size 1,
gradient accumulation of 4, learning rate $10^{-4}$, two epochs, a cosine
scheduler, a warmup ratio of 0.03, and random seed 42.

\begin{table}[ht]
\centering
\small
\caption{Recorded training and inference configuration.}
\label{tab:appendix_hyperparams}
\begin{tabular}{p{0.44\linewidth}p{0.36\linewidth}}
\toprule
\textbf{Setting} & \textbf{Recorded value} \\
\midrule
Base model & Qwen3.5-9B \\
Checkpoint source & \texttt{/home/models/Qwen3.5-9B} \\
Trainable modules & All linear modules \\
LoRA rank / scale / dropout & 8 / 32 / 0.05 \\
Precision & bfloat16 \\
Per-device batch size & 1 \\
Gradient accumulation & 4 \\
Learning rate & $10^{-4}$ \\
Epochs & 2 \\
Scheduler & Cosine \\
Warmup & 0.03 warmup ratio \\
Random seed & 42 \\
Maximum input length & 12{,}288 tokens \\
Maximum output length & 4{,}096 tokens \\
Temperature & 0 \\
Beam setting & One beam \\
Checkpoint selection & Final checkpoint after 2 epochs; preregistered global step 2160 \\
\bottomrule
\end{tabular}
\end{table}

\subsection{Serialization and Inference Contract}
\label{app:training:prompt}

Training examples require one JSON object with top-level keys
\texttt{annotation} and \texttt{rework\_plan}. The stable prompt contract is:

\begin{verbatim}
<System>
Return one JSON object with the declared annotation and rework_plan fields.
Do not add explanatory prose.
</System>
<User>
<public task representation>
...
</public task representation>
</User>
<Assistant>
{"annotation": ..., "rework_plan": ...}
</Assistant>
\end{verbatim}

During inference, output is parsed once. Raw output is recorded before
normalization and JSON parsing.
Contract, action-legality, and symbolic-executability failures are counted as
invalid; invalid actions are not counted as verified recovery.

\subsection{Baseline Protocol}
\label{app:training:baselines}

All baselines receive the same public task representation and shared output
schema under the same evaluation protocol.

\section{Evaluation Metrics}
\label{app:metrics}

We evaluate \textsc{FORGE} from four complementary perspectives:
structural sequence correctness, state and action parameter fidelity,
verified recovery, and hard-case rework quality. All automatic metrics are
computed from the structured prediction, the public symbolic transition
engine, and the reference annotation. Scores are normalized to $[0,1]$ and
reported as percentages in the main tables.

Let $\hat{D}_i$ and $D_i^{*}$ denote the predicted and reference diagnoses for
sample $i$, and let $\hat{\pi}_i$ and $\pi_i^{*}$ denote the predicted and
reference rework programs. The corresponding ordered action-identity
sequences are denoted by $\hat{A}_i$ and $A_i^{*}$. Let
$c_i$ and $c_i^{*}$ be the predicted and reference numbers of executable
actions. The symbolic replay predicate $R_i$ is true only when the predicted
output satisfies the schema, all actions are legal, the replay produces no
symbolic errors, and the final state matches the target state.

\subsection{Validity and Tolerances}
\label{app:metrics:validity}

A prediction is schema-valid when it parses under the declared output contract.
It is execution-valid when every action is legal in the state produced by its
predecessors, symbolic replay produces no error, and the terminal state is
equivalent to the target. Position, orientation, and scalar fields are
compared with the fixed tolerances used by the evaluator.

\begin{table}[ht]
\centering
\small
\caption{Evaluator comparison rules.}
\label{tab:appendix_tolerances}
\begin{tabular}{p{0.36\linewidth}p{0.42\linewidth}}
\toprule
\textbf{Field family} & \textbf{Recorded rule} \\
\midrule
Position / translation & Tolerance-aware comparison. \\
Orientation / rotation & Tolerance-aware comparison. \\
Scalar parameter & Tolerance-aware comparison. \\
Part identity & Exact identifier equality. \\
Action type & Exact action-name equality. \\
Target state & Pose, presence, and tolerance-aware symbolic comparison. \\
\bottomrule
\end{tabular}
\end{table}

\subsection{Structural and Parameter Metrics}
\label{app:metrics:structural}

\paragraph{Structural Sequence Exact (SSE).}
SSE measures whether the predicted ordered $(\textit{action},\textit{part})$
sequence matches the verified reference sequence exactly. It captures
structural sequence agreement, while action parameters and executable
restoration are evaluated separately by SAPF and DGTR:
\begin{equation}
\mathrm{SSE}
=
\operatorname{Macro}_{\mathrm{assembly}}
\left[
\frac{1}{|\mathcal{F}_a|}
\sum_{i\in\mathcal{F}_a}
\mathbf{1}\!\left[\hat A_i=A_i^*\right]
\right].
\label{eq:metric_sse}
\end{equation}
where $\mathcal{F}_a$ denotes the fault samples associated with assembly $a$.
The final score is macro-averaged across assemblies.

\paragraph{State--Action Parameter Fidelity (SAPF).}
SAPF evaluates the required observable diagnostic-state fields and the
reference-action parameters. Each field receives a binary score according to
the declared pose tolerance, rotation tolerance, or fixed scalar tolerance.
Missing fields receive zero credit. Let $s_i$ and $a_i$ denote the mean state
and action parameter fidelity for sample $i$:
\begin{equation}
\mathrm{SAPF}
=
\operatorname{Macro}_{\mathrm{assembly,type}}
\left[
\frac{1}{2}
\left(
\overline{s}_i+\overline{a}_i
\right)
\right].
\label{eq:metric_sapf}
\end{equation}
where each family is first averaged within a fault-type stratum and then
macro-averaged across fault types and assemblies. If one family has no
applicable fields for a sample, only the supported family contributes to the
sample score.

\subsection{Verified Recovery Metrics}
\label{app:metrics:verified}

\paragraph{Diagnosis-Grounded Target Restoration (DGTR).}
DGTR requires a valid output contract, exact fault diagnosis, successful
symbolic execution of the complete predicted program, and restoration of the
target state:
\begin{equation}
g_i
=
\mathbf{1}\!\left[
C_i
\land
(\hat D_i=D_i^*)
\land
R_i
\right],
\qquad
\mathrm{DGTR}
=
\operatorname{Macro}_{\mathrm{assembly,type}}
\left[g_i\right].
\label{eq:metric_dgtr}
\end{equation}
where $C_i$ indicates contract validity. No-fault samples are not included in
DGTR; they are handled separately through the empty-program requirement
defined by the output contract.

\paragraph{Verified Minimal Repair Efficiency (VMRE).}
VMRE measures how efficiently a verified prediction reaches the target relative
to the reference cost. A prediction receives efficiency credit only when its
diagnosis, execution, and required state and action parameters are valid:
\begin{equation}
\mathrm{VMRE}_i
=
\begin{cases}
\displaystyle
\min\!\left(1,\frac{c_i^*}{c_i}\right),
& g_i=1,\;v_i=1,\;c_i>0,\\[8pt]
0,
& \text{otherwise},
\end{cases}
\label{eq:metric_vmre}
\end{equation}
where $v_i$ indicates complete parameter validity. The reported VMRE is the
macro-average across fault types within each assembly, followed by an
assembly-level macro-average. Redundant actions reduce the score, while a
verified repair that matches or improves on the reference cost receives full
credit.

\paragraph{Minimal Verified Execution (MVE).}
MVE is a strict binary metric that combines diagnosis, execution, parameter
validity, structural sequence agreement, and reference-level action
efficiency:
\begin{equation}
m_i
=
\mathbf{1}\!\left[
g_i
\land
v_i
\land
(\hat A_i=A_i^*)
\land
(c_i\leq c_i^*)
\right],
\qquad
\mathrm{MVE}
=
\operatorname{Macro}_{\mathrm{assembly,type}}
\left[m_i\right].
\label{eq:metric_mve}
\end{equation}
MVE credits programs that reproduce the validated recovery sequence, restore
the target state, and use no more executable actions than the reference.

\subsection{Hard-Case Metric}
\label{app:metrics:hard}

\paragraph{Hard Reference Quality (HRQ).}
HRQ evaluates difficult fault cases whose metadata is marked as hard or expert.
A hard sample first must satisfy the diagnosis and verified-restoration gates.
Its remaining quality is determined by action-sequence similarity and supported
parameter fidelity. Let $d_i$ be the unit-cost edit distance between the
predicted and reference action-identity sequences:
\begin{equation}
q_i^{\mathrm{seq}}
=
\max\!\left(
0,
1-\frac{d_i}{\max(|\hat A_i|,|A_i^*|,1)}
\right).
\label{eq:metric_sequence_similarity}
\end{equation}

The hard-case score is then
\begin{equation}
\mathrm{HRQ}_i
=
\begin{cases}
\displaystyle
\frac{q_i^{\mathrm{seq}}+p_i}{2},
& g_i=1,\;h_i=1,\;p_i\ \text{is supported},\\[8pt]
q_i^{\mathrm{seq}},
& g_i=1,\;h_i=1,\;p_i\ \text{is unsupported},\\[4pt]
0,
& \text{otherwise},
\end{cases}
\label{eq:metric_hrq}
\end{equation}
where $h_i$ indicates membership in the hard subset and $p_i$ is the
state--action parameter fidelity for that sample. HRQ is macro-averaged across
assemblies and is reported only on hard fault samples. 
\subsection{Overall-Core and LLM-as-Judge}
\label{app:metrics:aggregate}

The main aggregate score is the unweighted arithmetic mean of the six
automatic metrics, computed from their unrounded values. A missing entry
shown as ``--'' in the result tables denotes zero credit and contributes zero
to this mean:
\begin{equation}
\mathrm{Overall\mbox{-}Core}
=\frac{\mathrm{SSE}+\mathrm{SAPF}+\mathrm{VMRE}+
\mathrm{DGTR}+\mathrm{MVE}+\mathrm{HRQ}}{6}.
\label{eq:overall_core}
\end{equation}

We additionally report an LLM-as-Judge semantic quality score on a fixed,
stratified 60-sample subset shared by all models. The judge assigns a score
from 0 to 5 based on diagnosis consistency, recovery completeness, action
executability, and target-state restoration. The judge score is reported
separately and is not included in \textsc{Overall-Core}.

Human evaluation provides an auxiliary comparison with the automatic metrics
and is not included in \textsc{Overall-Core}. The detailed evaluation protocol
is provided in Appendix~\ref{app:human_evaluation}.


\section{Error Analysis}
\label{app:error_analysis}

The automatic metrics summarize different aspects of executable rework quality,
but they do not show how a recovery attempt fails. We therefore organize the
qualitative analysis around the point at which a generated program departs from
the intended recovery trajectory. The resulting failure categories cover
diagnosis, dependency handling, action grounding, execution, and final
restoration.

\begin{table}[ht]
    \centering
    \small
    \setlength{\tabcolsep}{3pt}
    \renewcommand{\arraystretch}{0.95}

    \caption{Failure categories used in the qualitative analysis.}
    \label{tab:appendix_failure_taxonomy}

    \begin{tabular}{p{0.25\linewidth}p{0.49\linewidth}p{0.16\linewidth}}
        \toprule
        \textbf{Failure category}
        & \textbf{Description}
        & \textbf{Diagnostic signal} \\
        \midrule

        Fault misdiagnosis
        & The predicted fault type or affected part does not match the observed
        assembly state.
        & $\hat{D}_i \neq D_i^*$ \\

        Dependency omission
        & A required unlock, clearing, restoration, or relocking operation is
        missing from the recovery program.
        & Missing dependency \\

        Illegal grounding
        & The generated action or parameter is unavailable in the current
        symbolic state.
        & $\neg\operatorname{Legal}$ \\

        Parameter error
        & The action type is appropriate, but its pose, rotation, or other
        parameters fall outside the evaluator tolerance.
        & Invalid parameter \\

        Incomplete restoration
        & The program executes partially but the terminal state does not match
        the target configuration.
        & $S_T \not\equiv S^*$ \\

        Over-disassembly
        & The program moves additional components that are not required by the
        recovery route.
        & Excessive actions \\

        No-fault intervention
        & The model produces a non-empty program for an already valid assembly.
        & $\hat{\pi}_i \neq \varnothing$ \\

        Schema failure
        & The output cannot be parsed under the declared annotation and
        rework-plan schema.
        & $C_i=0$ \\

        \bottomrule
    \end{tabular}
\end{table}

\subsection{Failure Patterns in Executable Rework}
\label{app:rework_failure_patterns}

The failure categories expose different stages of executable rework
generation. Diagnosis errors occur when the model predicts an incorrect fault
type or affected component, which can subsequently produce an invalid recovery
structure.

Dependency errors arise when a required unlocking, clearing, restoration, or
relocking step is omitted. Such omissions can make an otherwise plausible
repair sequence impossible to execute under the assembly constraints.

Action-grounding errors are more local: the intended recovery step may be
correct, while the selected action, part identifier, pose, rotation, or other
parameter is invalid in the current symbolic state.

Finally, a program may execute partially but fail to restore the complete
target configuration, or may introduce unnecessary actions beyond the required
recovery route. These cases show how local errors can propagate into illegal
execution, incomplete restoration, or excessive intervention.

\section{Human Evaluation Protocol}
\label{app:human_evaluation}
Human evaluation uses the same fixed, stratified 60-sample subset as the
LLM-as-Judge evaluation. The subset is shared by all evaluated models so that
the automatic and human scores are computed on identical task instances.

Three experts with relevant experience in mechanical assembly, symbolic
planning, or structured-program evaluation independently rate every model
output in the subset. The experts use the same $0$--$5$ rubric as the LLM
judge, covering diagnosis consistency, recovery completeness, action
executability, and target-state restoration.

For each output, the three expert ratings are averaged with equal weight. The
resulting per-output scores are then averaged over the 60 sampled instances
to obtain a model-level human score. Human scores are reported separately and
are not included in \textsc{Overall-Core}.

\end{document}